\documentclass[12pt]{article}

\usepackage[utf8]{inputenc}
\usepackage{amsmath, amssymb, amsthm, mathtools, bbm, graphicx, fullpage, caption, subcaption, multirow, url, authblk, hyperref}
\usepackage[dvipsnames]{xcolor}
\usepackage[normalem]{ulem}
\usepackage{setspace}
\newcommand{\blkfootnote}[1]{\begingroup\renewcommand\thefootnote{}\footnote{#1}\addtocounter{footnote}{-1}\endgroup}

\title{Boosting the performance of a deep learning ECG classifier using an artificial model}

\author[1]{\small{Ismail Sadiq}}
\author[2]{\small{Erick A. Perez-Alday, PhD}}
\author[3]{\small{Amit J. Shah, MD}}
\author[2]{\small{Ali Bahrami Rad, PhD}}
\author[2]{\small{Reza Sameni, PhD}}
\author[2,4]{\small{Gari D. Clifford, DPhil}}

\affil[1]{\small{Department of Electrical \& Computer Engineering, Georgia Institute of Technology}}
\affil[2]{\small{Department of Biomedical Informatics, Emory University}}
\affil[3]{\small{Department of Epidemiology, Rollins School of Public Health, Emory University}}
\affil[4]{\small{Department of Biomedical Engineering, Georgia Institute of Technology \& Emory University}}

\date{10 December 2021}

\begin{document}

\maketitle

\blkfootnote{This work was partially funded by Emory University, 
and the National Institutes of Health/National Heart, Lung, and Blood Institute (awards R01HL136205, R01-HL109413-06, K23HL127251 and R03HL146879).
The content is solely the responsibility of the authors and does not necessarily represent the official views of the authors' sponsors and employers.

For information regarding this article, please contact the author via email. Address: Department of Biomedical Informatics, Emory University, Woodruff Memorial Research Building, 101 Woodruff Circle, 4th Floor East, Atlanta, GA 30322, USA. Phone: (404)-727-4631. Email: gari@gtech.edu
}

%
%

\newpage

\begin{abstract}

\textit{Objective}: Transfer learning has been shown to be an effective strategy for pre-training a deep neural network (DNN) to boost performance on smaller data-sets. This is particularly useful when it is difficult or costly to acquire high quality data, such as in medicine. In this study we evaluated the additional utility of pre-training on a computationally efficient but realistic physiological model which can be tuned to explore a wide variety of states for a specific condition. In this current work, we focused on the existence of T wave alternans (TWAs) as a marker of post traumatic stress disorder (PTSD) in a small cohort of 36  veteran twins.

\textit{Approach}:
Using a previously validated artificial ECG model, we generated 180,000 artificial ECGs with or without TWAs, with varying heart rate, breathing rate, TWA amplitude, and ECG morphology. We then took a previously developed state-of-the-art DNN, trained on over 70,000 patients to classify 25 different rhythms, modified the output layer to a binary class (TWA or no-TWA, or equivalently, PTSD or no-PTSD), and performed transfer learning on the artificial data. In a final transfer learning step, real the DNN was trained on ECGs from 12 individuals with PTSD and 24 controls, and evaluated using leave-one-subject-out cross-validation. The training and testing processes were repeated with and without each of the three data sets (rhythm data, artificial data, PTSD data) to evaluate the contribution of each transfer learning step. The area under receiver operating characteristic (AUROC) curve, accuracy (Acc), F1-score for PTSD and balanced accuracy (BAcc.) classification was reported for each trained model. 

\textit{Main results}: The best performing approach (AUROC = 0.77, Acc  = 0.72, F1-score = 0.64, BAcc. = 0.73) was found by performing both transfer learning steps, using the pre-trained arrhythmia DNN, the artificial data and the real PTSD-related ECG data. 
Removing the artificial data from training led to the largest drop in performance. Removing the arrhythmia data from training provided a modest, but significant drop in performance. 

\textit{Significance}: 
In healthcare, it is common to be resource limited, and only have a small collection of high-quality data. Moreover, many diseases are quite rare, and it is impracticable to assemble large databases amenable to machine learning. Conversely, large databases are hard to curate, and often the quality of the labels drops as the volume increases. Here we presented a solution to these issues through transfer learning on a large realistic artificial database. By tuning the artificial model to generate data which closely matched the pathologies expressed in the target population, while allowing the model to explore a vast range of both normal and pathological morphologies, we created a novel (artificial) training database that significantly boosted the performance of the classifier. Finally, it is worth noting that a DNN trained on arrhythmia data only can outperform traditional methods for identifying cardiac signatures of PTSD, which reinforces the idea that PTSD is strongly associated with arrhythmogenesis. 

The paradigm presented here, involving model-based performance boosting, is likely to be generalizable to other pathologies, and potentially to other data modalities, particularly within medicine and biology. Given that data sourcing, transfer and preservation are costly, and compute is relatively cheap, our approach could have enormous potential in the biological sciences.


\end{abstract}

\section{Introduction}
\label{sec:introduction}

As we continue to assemble, and publish increasingly larger collections of biomedical data, the quality of data tended to drop
\cite{Clifford2009,10.1145/3411764.3445518}. Perhaps more importantly, the labels associated with the data also drop in quality as the database size increases. These issues arise because the resources required to hand curate the data and labels scale with the data volume
\cite{DBLP:journals/corr/abs-2007-10502}. Moreover, larger datasets were often collected serendipitously, or for other reasons (e.g., electronic medical records, or other non-research medical data). Data are therefore not coded or labelled appropriately for the scientific question that is being asked by a researcher at a later date. 

One potential solution to this issue is the use of transfer learning 
\cite{8583799,9498418,LIANG2020104964}
and domain adaptation \cite{5640675} to pre-train models on  more ``trustworthy'' data. Of course, this may ``bake-in'' the biases of the original data, which tend to be skewed towards the well-funded researchers that collected the original data. In particular, the bias is often away from people of color, women, and other minorities, particularly those in Low and Middle Income Countries (LMICs) \cite{10.1093/jamia/ocaa094,10.3389/frai.2020.561802}. 

While we, and others, have demonstrated the utility of transfer learning and domain adaptation to leverage large biomedical databases for new tasks, ranging from sleep staging to eye tracking \cite{9144416,9360436,9079571,Nasiri2020,Li2021,10.1093/sleep/zsab027}, to the best of our knowledge, no published work has yet to emerge which incorporates the potential of  physiological models to boost the performance of a learning algorithm. That is, if a suitable model existed from which we were able to generate new ``realistic'' patient data, over a range of conditions, we could significantly augment a small database of under-represented individuals, with similar data, simultaneously improving performance and reducing bias towards the small and relatively homogenous 
source data.

In this work, we used a well-known computationally efficient yet realistic model of the electrocardiogram (ECG) to generate a large dataset (over 180,000) ``recordings'' of a subtle cardiac abnormality largely unrelated to the arrhythmia database (of over 70,000 subjects) on which a deep convolutional neural network (DCNN) was trained and validated. Using transfer learning, the network was adapted to the artificial ECG, and then to a much smaller database of only 36 individuals to test the hypothesis that realistic artificial data can significantly boost the performance of a DNN on databases comprising limited population numbers.  

\section{Background}

\subsection{Synthetic data}
A closely related area of research to the work proposed here is synthetic data generation, which is receiving increasing attention of late, is that of synthetic data generation. The majority of articles appear to have focused on synthetic medical data generation have focused on imaging and electronic medical records
\cite{Goncalves2020,
Dube2014,
Kim2015,
Buczak2010,
Walonoski2018,
Chen2019,
Sankaranarayanan2018,
Yang2018,
Zhang2019,
Tucker2020,
Chen2021,
Mahmood2018,
Tang2021,
Yale2020,
Nie2018,
Ive2020,
Prakosa2012,
Moniz2009,
Benaim2020}.
In fact, examples of synthetic data generation to improve classifier performance are too numerous to effectively review, and are not the focus of this research. While synthetic data are useful for boosting data representation from minority classes, or improving the robustness of applying data driven approaches such as machine learning to a variety of problems, there are inherent dangers. First, there are few guarantees that the artificial data are generated in a manner consistent with reality. The choices of the researcher largely drive the distribution of augmentation. 
More recent work in the use of generative artificial neural networks has `automated' these choices somewhat, but at the cost of lacking generalization outside the distribution of the source data. Conversely, a model-based approach provides for the possibility of generating realistic (and physiologically bounded) data outside the observation distribution.

\subsection{Artificial physiological data}
Again, a survey of artificial physiological data generation is outside the scope of this article. The reader is referred to several recent surveys on the selection, utility and validity of physiological models 
\cite{Kerkhof1999,
Pruett2020,
Bighamian2021}. The key point is that clinicians, working with experts in machine learning, find that a model-based approach is still of utility in medicine
\cite{Sarma2020}. Even though big data/deep learning approaches are starting to outperform domain-expert driven approaches to feature creation, clinical teams have a deep distrust of brute-force data-driven models. Ultimately, a clinician finds it hard to be comfortable with a black box (lacking explanation of the reasons for the inclusion of the certain features).   
Model-based approaches allow the development team to test a wide variety of possible `patients', particularly edge cases or rare pathological scenarios. Such stress tests of the code are critical for building confidence. While it is possible to claim  that \emph{eventually} clinicians will come to accept modern machine learning in the same way that they have accepted other key black box technology (such as bedside arrhythmia alarms).  However, acceptance is much more likely if the developed algorithms allow some form of reasoning that map back to traditional (compartmental) physiological models.

In this work we propose one further use of artificial models - to generate (pre-) training data for deep neural networks, in order to learn specifics of a rare pathology (or a pathology that is expensive and time-consuming to capture). 

\subsection{Clinical domain example}

The example medical condition on which we  demonstrate the utility of the proposed approach was chosen to be post-traumatic stress disorder (PTSD), since data on PTSD patients is difficult to collect.
PTSD is a psychiatric disorder occurring in individuals who had experienced/witnessed terrifying events for a prolonged duration. These events included but were not limited to wars, physical and sexual abuse, traffic accidents, or natural disasters like earthquakes. Symptoms included negative thoughts, social distancing, and avoiding places, activities, and people that served as reminders of the event. In most cases, the individuals recovered from the experience in a few weeks or months, needing support from family and friends. Some cases were more severe and required therapeutic intervention.

In the United States, war veterans, in particular, were observed to develop symptoms and treatment costs were USD\$8300 on average per veteran annually \cite{ptsdexpense}. Early detection could lead to an early start in the recovery cycle, delayed treatment of the disorder would prolong suffering and may reduce treatment efficacy \cite{Maguen2014ptsdtreatment}. Delaying treatment may also cause the symptoms to become worse and result in a poor quality of life \cite{delayptsdtreat}.

In hospitals, PTSD assessments are performed as structured clinical interviews by experienced healthcare workers. The clinician-administered PTSD scale (CAPS-5) is the gold standard for PTSD diagnosis and consists of a 30-item questionnaire that can take from 45-60 minutes to complete \cite{Jiang2020}. The CAPS-5 can be used to make a current PTSD diagnosis (past month), lifetime PTSD diagnosis (worst month) or a PTSD assessment over the past week, depending on the period for which the individual is examined. The response to each item on the questionnaire is noted as a severity score on a scale of 0-4. Zero corresponds to the absence of symptoms and four corresponds to the highest level of severity. The responses in the interview are highly subjective and can vary depending on the openness or consistency of the subject and the skill of the clinician/rater. The overall process for PTSD assessment is time-consuming and burdensome on the healthcare workers, aside from the existence of 
variability in the subject's responses or rater's skill in administering the survey. 
Improving the screening test for PTSD with machine learning would reduce the burden on the healthcare providers and shorten the latency in the time to treatment for individuals who had PTSD.

Previous studies have trained machine learning algorithms to 
classify PTSD status from the responses for self-reported PTSD surveys. Wshah \textit{et al.}~\cite{Wshah2019} trained an ensemble of classifiers on a reduced set of responses from the PTSD checklist-5 (PCL-5) that were collected through smartphones. Ilhan \textit{et al.}~\cite{Ilhan2015} used a similar approach of training a classifier on responses to a questionnaire similar to the PCL-5. The authors used feature selection to remove uninformative questions, which improved classification performance. Jiang \textit{et al.}~\cite{Jiang2020} trained a random forest classifier on the responses to the self-administered interview for the diagnostic statistical manual for mental disorders (DSM-5). They achieved an accuracy, a sensitivity, a specificity, a positive predictive value (PPV), a negative predictive value (NPV) and an area under receiver operating characteristic (AUROC) curve, all above 0.9 using the top 14 response features determined using the Gini impurity \cite{Breiman1984gini}. Their results suggested that structured interviews for PTSD screening could be abbreviated without losing accuracy, thus reducing the burden on clinicians. 
The 
inter-subject variability 
in the responses 
to the questionnaire and inter-rater variability 
remained high, however. Other more objective features have been sought. Marmar \textit{et al.}~\cite{Marmar2019} extracted features from speech recorded from a group of veterans with or without PTSD to determine PTSD status. Speech signals were easily recorded and transmitted. The 18 most important features for PTSD classification were determined through `shaving' and used to train a random forest classifier. Schultebrauks \textit{et al.}~\cite{schultebraucks2020} trained a deep belief network (DBN) on features extracted from audio and video recorded for individuals who had undergone a traumatic event and were being evaluated for PTSD.

With respect to cardiovascular data,  Reinertsen \textit{et al.}~\cite{Reinertsen2017ptsd} demonstrated that heart rate variability, recorded at the cardiovascular nadir (during deepest sleep) could identify PTSD withing a cohort of 72 subjects with an AUC of 0.86. Cakmak \textit{et al.}~\cite{Cakmak_9333654} demonstrated that circadian rhythm changes, measured by a wrist-worn research watch are predictive of post-trauma outcomes in a cohort of 1618 post-trauma patients. The highest cross-validated performance of research watch-based features (derived from pulse and accelerometer) was achieved for classifying participants with pain interference (AUC=0.70). A survey-based model achieved an AUC of 0.77, and the fusion of research watch features and ED survey metrics improved the AUC to 0.79.

In this study,  we trained an end-to-end DCNN on multi-lead ECG data to classify PTSD in order to demonstrate the strength of the association of myocardial electrical disturbances with PTSD, and to identify whether an appropriate electro-physiological model of the heart can provide a significant boost in classifier performance. Our aim was not to surpass  state-of-the-art results (because there is no public PTSD database with sufficient ECG), but merely to identify to what degree electrophysiology is connected with PTSD beyond the current traditional markers (and how we can optimally select data or models to identify subtle markers of PTSD). 
The motivation for this is that PTSD has been shown to be associated with arrhythmogenesis \cite{Vaccarino2013ptsd} and increased T wave alternans (TWA) \cite{LAMPERT20151000}. 
However, our target population consisted of only 36 individuals. To prevent the model from overfitting on this small dataset, and to improve generalization, the DCNN was pre-trained on over 70,000 ECGs taken from the PhysioNet Challenge 2021, as well as a novel dataset of artificial ECGs with varying levels of TWA amplitudes, generated specifically for this work.


\section{Methods}

\subsection{Data}
\label{ssec:data}
\begin{figure}[tbh]
         \centering
         \includegraphics[width=\textwidth]{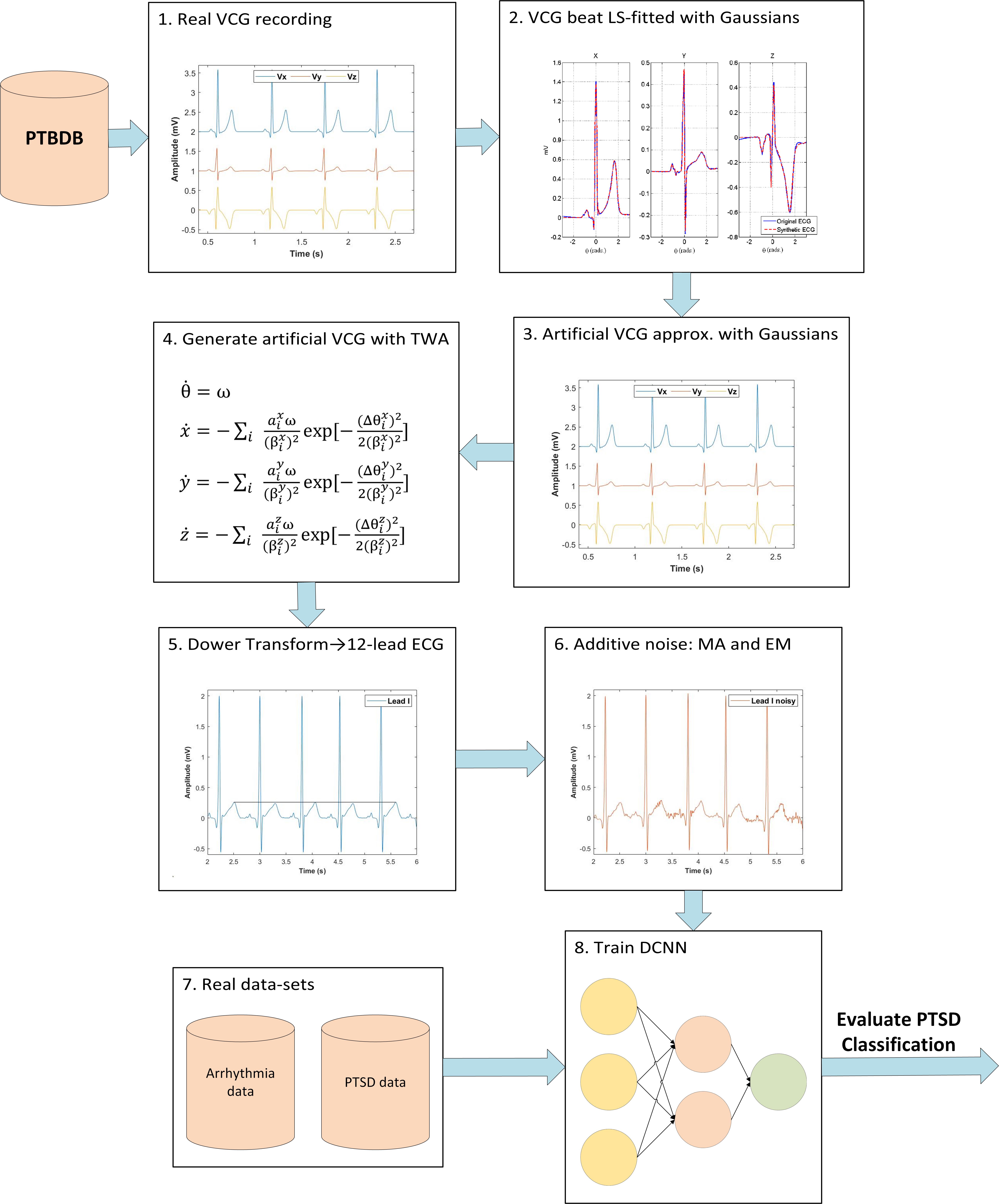}
         \caption{}
     \caption{A flowchart illustrating the process of generation of artificial ECG with TWAs. Steps 1a and 1b: 47 real VCG records were acquired from the PTBDB on PhysioNet. Step 2: Least-square estimates for Gaussians were derived for the average beat in the X, Y and Z VCG leads for each record. Steps 3 and 4: The derived Gaussian estimates were used for generating the artificial VCG with TWAs. Step 5: The Dower transform was applied to the VCG to obtain 12 ECG, lead I was used in the subsequent analysis, Step 6: Muscle artifact (MA) and electrode movement (EM) noise from the NSTDB on PhysioNet were added to the artificial ECG before training the classifier so it would be robust to noise. Steps 7 and 8: The artificial TWA data was used in combination with the real arrhythmia data and PTSD data to train a deep convolutional neural network (DCNN) for PTSD classification.
     }
        \label{fig:artifecg_pipeline}
\end{figure}

Three data sets were used to train the 
DCNN for a binary task of classifying a subject as having PTSD or not.
Figure \ref{fig:artifecg_pipeline} summarizes the generation of the artificial ECG with TWAs, Section \ref{ssec:artif_ecg}, and the training of the DCNN using each of the 3 data-sets explained in the following sections.

\subsubsection{Arrhythmia Data: The PhysioNet/ Computing in Cardiology (CinC) Challenge 2021}
The first data-set used in this work (to pre-train the DNN) is the twelve-lead ECGs from the PhysioNet/CinC Challenge 2021 ECG database, which  consists of over 70,000 labeled twelve-lead ECGs collected from 6 different hospitals across 3 continents \cite{Reyna_2021b,Perez_Alday_2021,Physionet}. 
From hereon, the PhysioNet/CinC Challenge 2021 ECG data are referred to as the real \emph{arrhythmia data}.
The recordings exhibit one or more of twenty-five labeled different arrhythmias, including atrial arrhythmias, ventricular arrhythmias and normal sinus rhythm. Since PTSD is associated with arrhythmogenesis \cite{Rosman2019}, we expect hat arrhythmia data will have some relationship with the PTSD label, which is discoverable by a deep neural network. 

\subsubsection{Model-based artificial ECG}
\label{ssec:artif_ecg}
The second data-set consists of artificial ECG generated with varying amplitudes of TWA. Similar to Clifford \textit{et al.}~\cite{Clifford2006artifecga,Clifford2005artifecgb,Clifford2007artifecgc,Clifford2010}, the morphologies of these artificial ECGs were derived using a least-square fit of Gaussian parameters (in the VCG representation) to normal subjects in the Physikalisch-Technische Bundesanstalt database (PTBDB) \cite{BousseljotKreiselerSchnabel+1995+317+318}. The intuition behind this approach was that any continuous function 
can be approximated arbitrarily well with a finite number of Gaussian functions \cite{BenArie95,McSharry2003}. The average beats for each subject were estimated from the VCG recordings in the [X, Y, Z] orthogonal vector directions. Gaussian functions with varying amplitude and standard deviations were placed at different instants during the beat's cardiac cycle to best approximate the average beat morphology in each VCG recording by minimizing the squared error. 
The functionality for the least-square estimate of Gaussian parameters fitted to a VCG beat was implemented as part of the open-source electrophysiological toolbox \cite{SameniOSET}. The Dower transform was then applied to the artificial VCG to generate the twelve-lead ECG representation \cite{Dower1980}. The ECG simulator could accurately generate abnormalities such as TWAs of varying amplitudes. For details on the generation of the artificial ECG with TWAs refer to \cite{Clifford2010} and \cite{Sadiq_2021}. The code for the ECG simulator is available as part of an open-source toolbox \cite{Vest2017Cardiotool}. 

Elevated TWAs 
were associated with elevated levels of stress by Lampert \textit{et al.}~\cite{LAMPERT20151000}. However, no digital database of significant size existed for ECGs labeled with TWAs. For the purpose of this study we generated artificial ECGs with or without TWAs. The morphologies of the artificial ECGs were derived from 47 normal subjects in the PTBDB. The HRs for the ECGs was varied from 60-110 beats per minute (bpm) in increments of 2, the upper limit for the HRs was kept at 110\,bpm instead of 100\,bpm since individuals with PTSD were known to have slightly elevated HRs than normal. The BR was varied between 12-20 respirations per minute (rpm) in increments of 1. For ECGs with TWAs, the TWA amplitude was varied between 20 to 100\,$\mu$V in increments of 1\,$\mu$V. To augment the artificial ECG data random perturbations were added to the Gaussian parameters for amplitude and standard deviation, used for estimating the ECG morphology in a recording. For each of the 47 subjects used, a maximum of $\pm$4.5\% percent of the parameter value was used to set the limits of a uniform distribution centered at the original parameter value and a number uniformly sampled from this distribution was assigned to the respective parameter. 

The limit of $\pm$4.5\% was determined empirically. For each of the 47 subjects the range of HRs were determined for which the QTc interval remained within the normal physiological range of between 360\,ms to 440\,ms \cite{Clifford:2006:AMT:1213221}. The QT interval was corrected using the `Bazett' correction~\cite{bazett1920analysis}. Next, the percentage of the value of the amplitude or standard deviation parameter used to limit the distribution of perturbations was increased from 1\% to 10\% in increments of 1\% before fine tuning. The perturbation was uniformly sampled from each distribution. For a maximum perturbation of at most $\pm$4.5\% of the parameter value, more than 95\% of the QTc intervals were within the normal range specified above. Therefore perturbations were added within the $\pm$4.5\% of the parameter value range. The additional check on the QTc interval was added so the artificial ECGs generated with perturbations would resemble ECGs from normal individuals with realistic characteristics. The model-based artificial ECG data was referred to as artificial TWA data. In addition electrode movement 
and muscle artifact 
noise from the noise stress test database on PhysioNet was added to train the model to be robust to noise for significant TWA detection \cite{Moody_1984}. Electrode movement and muscle artifact noise was added in equal proportion to each ECG window for a signal to noise ratio (SNR) between 15 to 30 dB. Baseline wander 
was not added as it was easily removed through median filtering. A total of 180,000 artificial ECGs were generated, half of which had TWAs and half did not. For pre-training the DCNN, the artificial ECGs were divided into 10 folds in a stratified manner. Eight folds were used for training, 1 fold for validation and 1 for testing. The weights with the lowest error on the validation fold were used for evaluating the test set classification accuracy. The network was trained as described in Section \ref{ssec:dnn}.

\subsubsection{PTSD data and preprocessing}
The PTSD data consisted of 12 subjects with PTSD and 24 controls. The data was sampled at 1000 Hz, consistent with previous algorithms that used deep neural networks trained on ECG data to detect cardiac arrhythmias 
\cite{Fu2020}. The data consisted of single channel Lead I ECG and the amplitude resolution of the data was measured to be 1.32$\times$10\textsuperscript{-6} mV. Hence the data was high resolution both in the temporal and spatial domains. 

Each subjects recording in the PTSD data-set was segmented into 16-second windows with 20\% overlap. Fifty 16-second windows with a signal quality index (SQI) of 1 were randomly selected for each individual. The signal quality index by Li \textit{et al.}~\cite{Li2014sqi} was used to identify clean ECG analysis windows that were minimally affected by noise \cite{Sadiq_2021}. The algorithm used two sets of fiducial point detectors, one being robust to noise and the second being sensitive to noise. When the fiducial points detected by either detector were consistent the signal was considered to be clean, in case of disagreement between the detections the signal was considered noisy. Baseline wander was removed using the 2 stage median filter by de Chazal \textit{et al.}~\cite{deChazal2004}.

\subsection{Deep convolutional neural network architecture and training}
\label{ssec:dnn}

\subsubsection{Network architecture}
\label{sssec:mdl}
\begin{figure}[tbh]
     \centering
     \begin{subfigure}[b]{0.4\textwidth}
         \centering
         \includegraphics[width=0.55\textwidth]{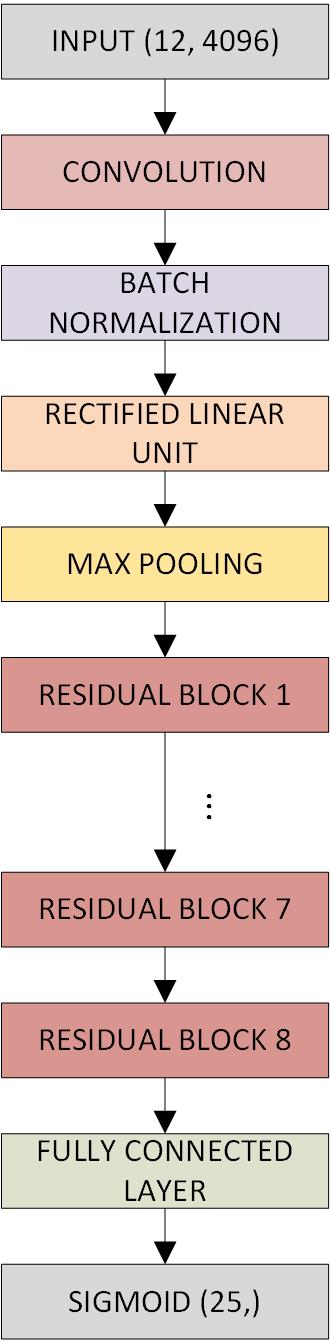}
         \caption{}
     \end{subfigure}
     \hfill
     \begin{subfigure}[b]{0.4\textwidth}
         \centering
         \includegraphics[width=0.55\textwidth]{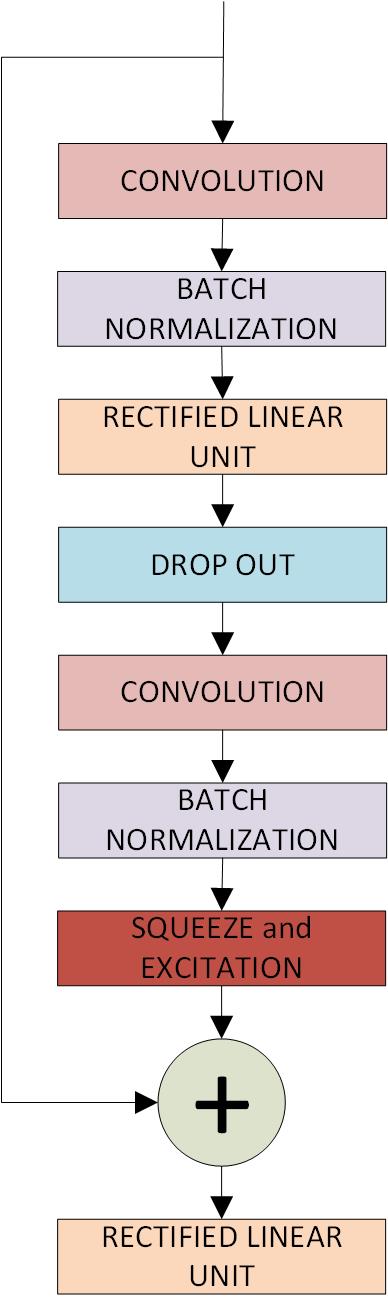}
         \caption{
         }
     \end{subfigure}
     \caption{(a) ResNet model pre-trained on the real arrhythmia data and used for classifying as PTSD. (b) Residual block.
     }
        \label{fig:resnet}
\end{figure}

The 
DCNN used was the same as that developed by 
Zhao \textit{et al.}~\cite{Zhao2020}. The network was selected as overall it was the second best algorithm at the PhysioNet challenge in 2020. In addition the network was the top performing algorithm for the category of algorithms that learned all the features from the input ECG and did not require evaluation of any additional hand-crafted features to be presented as an input. The architecture is 
given in figure \ref{fig:resnet}a.

The following models were trained to classify the PTSD ECG data using the data-sets described in Section \ref{ssec:data}.

\begin{figure}[tbh]
\centering 
\begin{subfigure}{0.205\textwidth}
  \includegraphics[width=\linewidth]{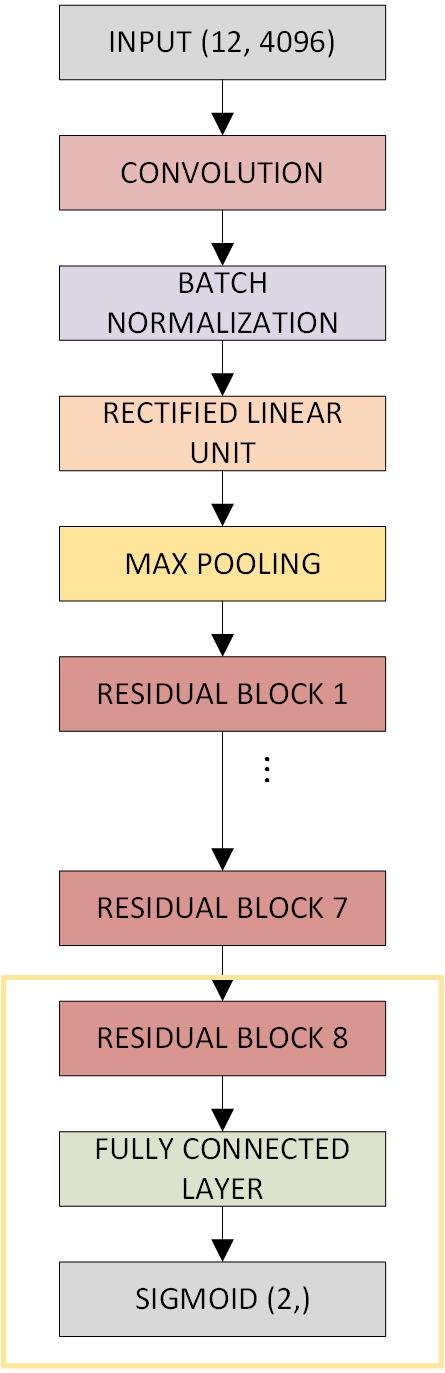}
  \caption{ResNet model 1}
\end{subfigure}\hfil 
\begin{subfigure}{0.2\textwidth}
  \includegraphics[width=\linewidth]{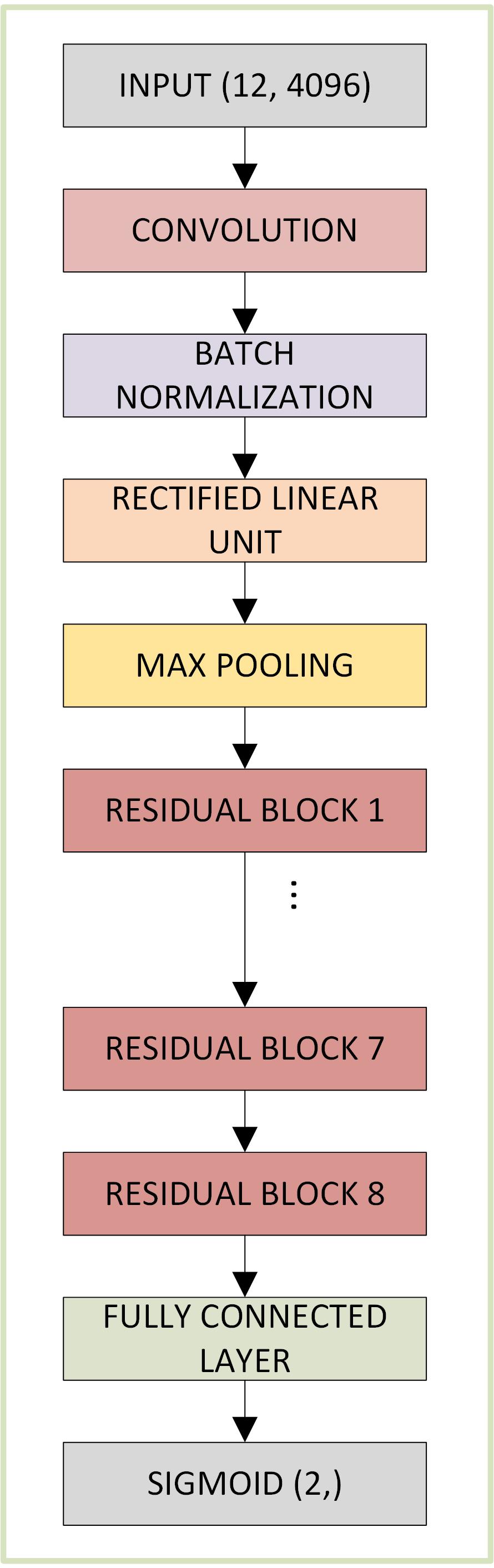}
  \caption{ResNet model 2}
\end{subfigure}\hfil 
\begin{subfigure}{0.2\textwidth}
  \includegraphics[width=\linewidth]{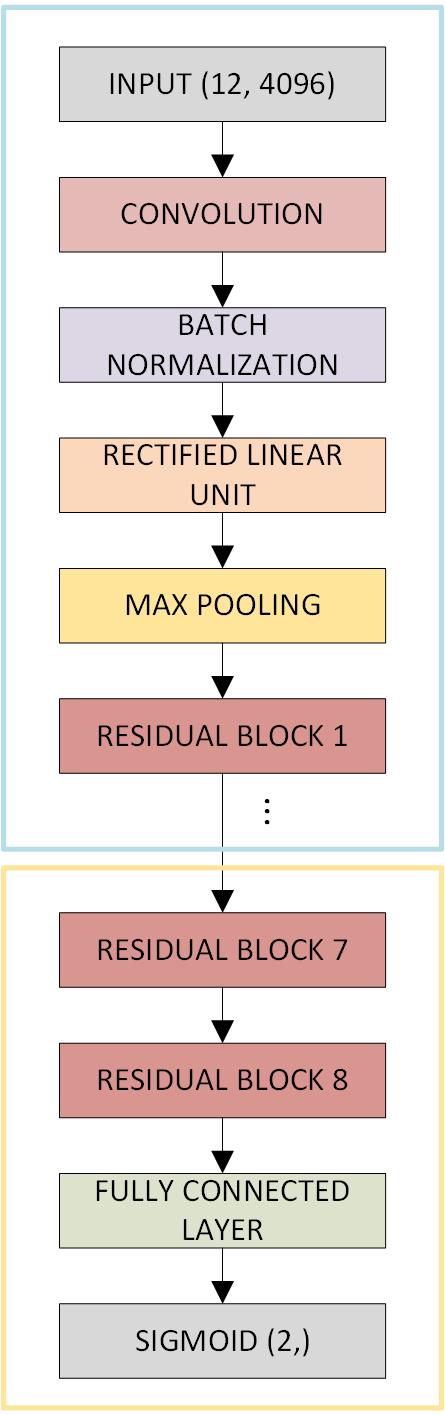}
  \caption{ResNet model 3}
\end{subfigure}

\medskip
\begin{subfigure}{0.2\textwidth}
  \includegraphics[width=\linewidth]{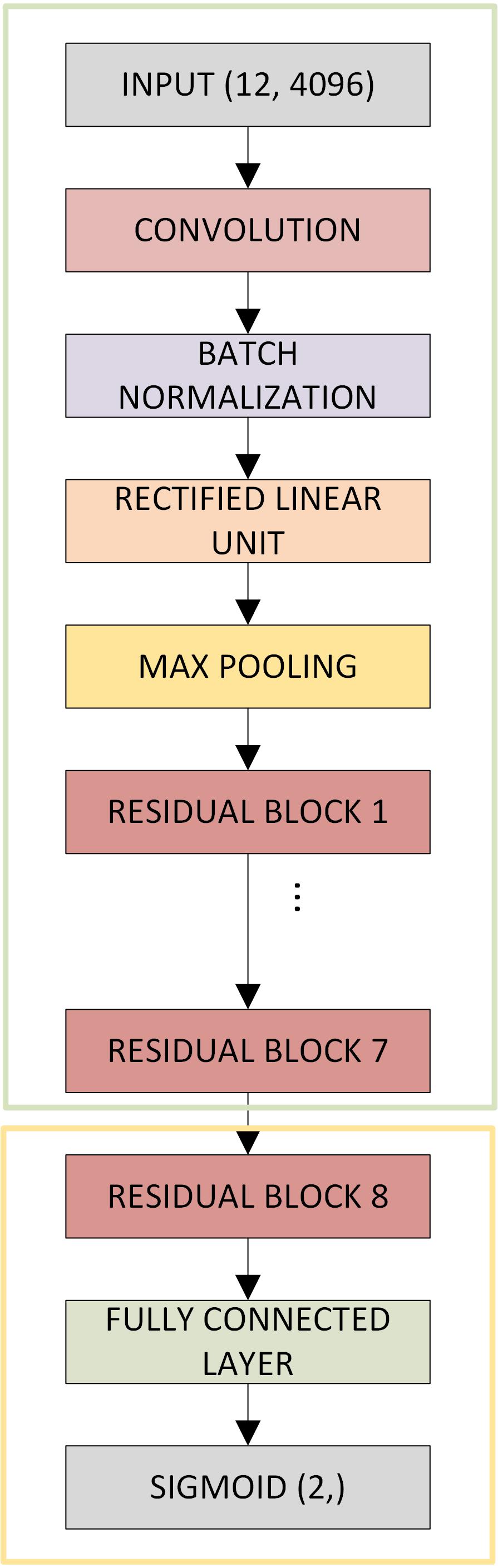}
  \caption{ResNet model 4}
\end{subfigure}\hfil
\begin{subfigure}{0.2\textwidth}
  \includegraphics[width=\linewidth]{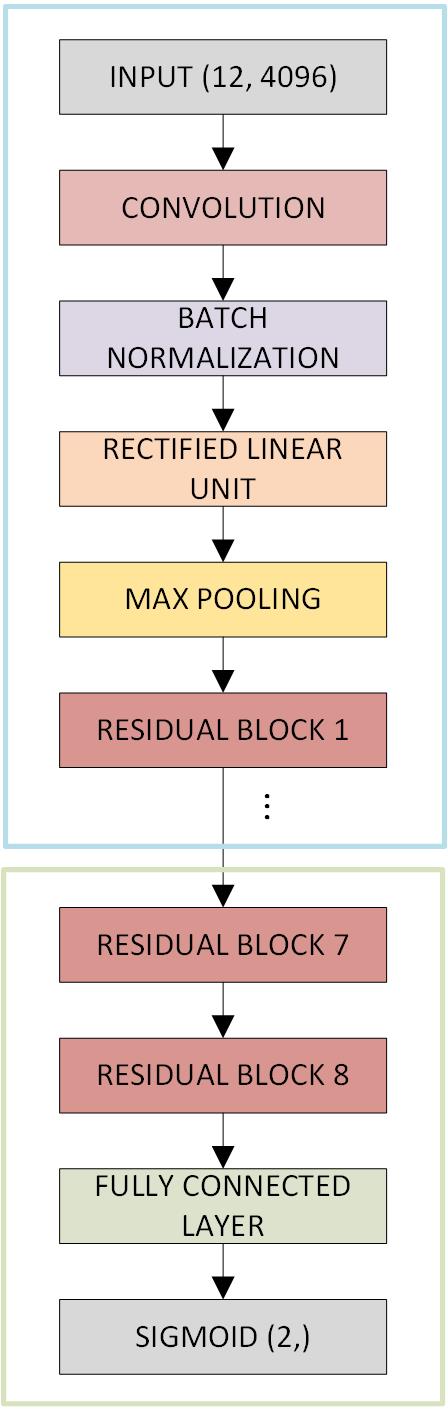}
  \caption{ResNet model 5}
\end{subfigure}\hfil
\begin{subfigure}{0.2\textwidth}
  \includegraphics[width=\linewidth]{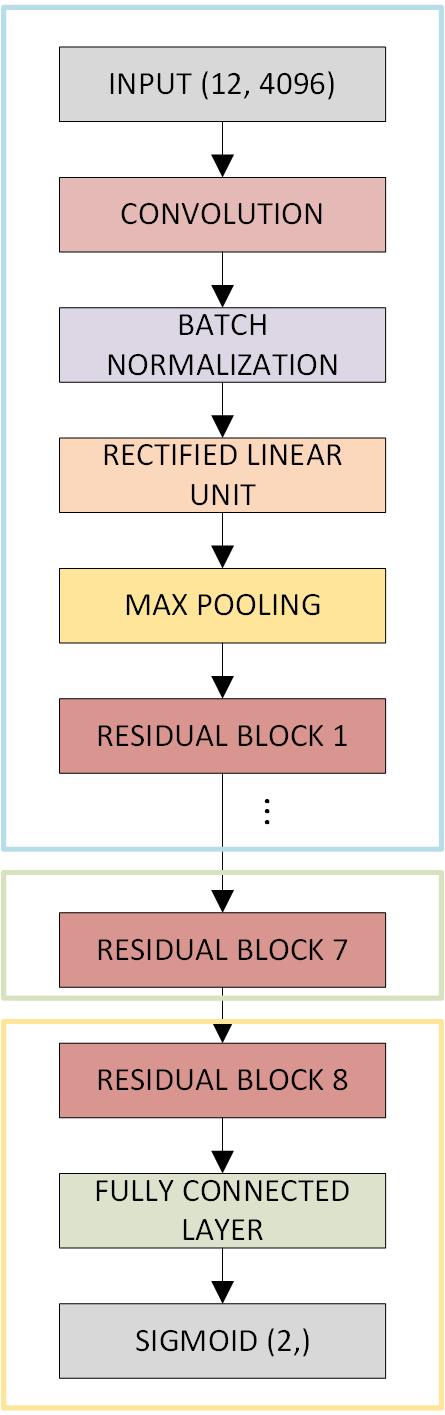}
  \caption{ResNet model 6}
\end{subfigure}

\caption{How each deep convolutional neural network model was trained is illustrated. The layers of the network trained on the real arrhythmia data are enclosed in blue. The layers of the network trained on the artificial TWA data are enclosed in green. The layers of the network trained on the PTSD data are enclosed in yellow.}
\label{fig:three graphs}
\end{figure}

\begin{itemize}
\item \textbf{Baseline (BL) model}: Hand-crafted features and logistic regression were used as the baseline against which deep learning approaches were compared. A logistic regression classifier was trained on features extracted from TWAs measured using the modified moving average (MMA) method. To reject noise-triggered TWA, a surrogate statistical test \cite{Nemati2011twa}, provided as part of the open source PhysioNet Cardiovascular Signal Processing Toolbox \cite{Vest2017Cardiotool} was used. The TWAs were measured in 60 beat windows with 50\% overlap. TWAs were considered statistically significant compared to the noise threshold from the surrogate test with a p-value $\leq 0.05$. The significant TWA detections were divided by HR decile into the following bins: [30,60), [60,70), [70,80), [80,90), [90,100) and [100,110], measured in bpm. For each individual the mean significant TWA amplitude was computed in each bin and used as a feature vector. The AUROC curve was computed for leave one (subject) out testing, i.e. one subject was held out for testing while the features from the remaining subjects were used for training the logistic regression classifier. The process of training and testing the classifier was repeated so that each subject was considered as the test subject. 
\textbf{Only the target PTSD data was used for training the logistic regression classifier.}

\item \textbf{Model 1:} The weights of the ResNet model based on Zhao's approach were randomly initialized. Only the 8\textsuperscript{th} residual block (ResB) and the fully connected (FC) layer weights were trained using the PTSD data. Training the 8\textsuperscript{th} ResB and FC layer allowed direct comparison with model 6 to determine if pre-training of the layers before the 8\textsuperscript{th} ResB 
with the 
real arrhythmia data and the artificial TWA data improved classification performance on the target PTSD data. \textbf{No real arrhythmia data or artificial TWA data were used for transfer learning; only the target PTSD data were used to train and test.}

\item \textbf{Model 2:} The weights of the ResNet model based on Zhao's approach were randomly initialized and the complete model was trained on the artificial TWA data. The trained classifier 
was used to directly classify the PTSD data. \textbf{No real arrhythmia data or PTSD data were used for training the model.}

\item \textbf{Model 3:} The weights of the ResNet architecture were pre-trained on the 
real arrhythmia 
data, after which transfer learning was performed by training the 7\textsuperscript{th} ResB, 8\textsuperscript{th} ResB and the FC layer using the PTSD data only. \textbf{No artificial TWA data was used for pre-training.}

\item \textbf{Model 4:} The weights of the ResNet model based on Zhao's approach were randomly initialized and the model was trained on the artificial TWA data. A transfer learning step was then performed to train only the 
8\textsuperscript{th} ResB and the FC layer using the PTSD data. \textbf{No real arrhythmia data was used for pre-training.}

\item \textbf{Model 5:} The weights of the ResNet architecture were pre-trained on the real arrhythmia data. Transfer learning was performed on the 7\textsuperscript{th} ResB, 8\textsuperscript{th} ResB and the FC layer with the artificial TWA data for classifying TWA presence. \textbf{No PTSD data was used - only pre-training with the real arrhythmia data and the artificial TWA data were used with a transfer learning step.}

\item \textbf{Model 6:} The weights of the ResNet architecture were pre-trained on the real arrhythmia data. Transfer learning was performed on the 7\textsuperscript{th} ResB , 8\textsuperscript{th} ResB and FC layer using the artificial TWA data. A second transfer learning step was then performed to train only the 8\textsuperscript{th} ResB and FC layer using the PTSD data. \textbf{All data-sets were used and two transfer learning steps were used.}
\end{itemize}
Figure \ref{fig:three graphs} illustrates 
the data-sets used for training each DCNN model.
The network layers enclosed in blue were trained on the 
real arrhythmia data, the network layers enclosed in green were pre-trained on the artificial TWA data and the network layers enclosed in yellow were trained on the PTSD data. In model 6, the first transfer learning step on the artificial TWA data was considered a stepping stone before performing transfer learning on the PTSD data. If model 6 outperformed model 4, additionally training on the artificial TWA data was beneficial for PTSD classification. Evaluating the performance of model 1 would determine if pre-training on the real arrhythmia data and artificial TWA data was important for accurate PTSD classification. Model 5 was completely trained on the real arrhythmia data after which transfer learning was performed on the 7\textsuperscript{th} ResB, 8\textsuperscript{th} ResB and the FC layer, with the artificial TWA data. The trained model 5 was used to classify the PTSD data, to determine if transfer learning on the PTSD data was an important step to correctly classify subjects as suffering from PTSD or not.  


\subsubsection{Network training}
The ResNet architecture was trained for 50 epochs with an ADAM optimizer. The learning rate was initially set to 0.003 and was reduced by a factor of 0.1 after the 20th and 40th epoch. The batch size was kept at 64. After pre-training on the real arrhythmia data the final sigmoid function layer was replaced to have 2 outputs. When classifying the artificial 
TWA data and PTSD data all the input leads except lead I were reduced to zero.
To account for the class imbalance when training on the PTSD data, 25 of the available 50 windows for each control subject not being tested, were used to constitute the training data.

\subsection{Evaluation methodology: Leave one out testing}
\label{sssec:loot}
Given the limited number of subjects in the PTSD data-set, the performance was evaluated on unseen data with the following leave-one-out testing (LOOT) approach. One subject was held out as the test subject and was not used for training or validation. The remaining subjects were used for training and validation. Ten percent of each subject's data in the training set was used for validation. The classifier was trained until the validation loss was observed to increase for three consecutive cycles. The classifier with the lowest error on the validation data was used for classifying the test subject. The procedure was repeated to test each subject and the fraction of windows for each subject classified as PTSD was used to compute the AUROC curve. The accuracy, balanced accuracy and F1 score for the optimal operating point on the receiver operating characteristic (ROC) curve was computed for each model. Due to class imbalance in the PTSD data the balanced accuracy was computed as the un-weighted average of sensitivity and specificity. The optimal operating point was computed as follows, a line with a slope of 1 started at the top left corner of the plot and was lowered until it intersected the ROC curve, the point at which the line first intersected was considered the optimal operating point. Defining the slope to be 1 resulted in the equal importance of the sensitivity and specificity, so the determination of the optimal point on the ROC curve was not affected by the imbalance in classes.

\subsubsection{Evaluating trained models on artificial TWA data}
Each of the models trained during LOOT was used to evaluate the classification performance on the artificial TWA data. This step was performed to determine wehether the models could still generalize back to. larger cohort, or whether they had overfit on the relatively small numbers of PTSD subjects. The mean and variance for the AUROC curve, sensitivity and specificity with LOOT were reported for models 1-6 and the baseline model mentioned in section \ref{sssec:mdl}. A $\chi^2$ test was performed between the classification labels for the artificial TWA data from model 6 and the true labels.

\subsubsection{Leave one out cross validation}
\label{sssec:loocv}
The classification achieved through LOOT was expected to be lower than the classification performance achieved given a larger number of data samples. The training and validation data sets consisted of ECG windows from the same individuals, therefore the classifier would overfit to particular trends in the ECGs for the individuals forming the training and validation data. The trained classifier was therefore not expected to generalize as well to ECG data from unseen individuals.
In addition, the performance for the best model with LOOT with respect to AUROC curve was evaluated using leave one out cross validation (LOOCV). The classification achieved through this approach was an optimistic estimate since it was not tested on unseen data and validation set loss is usually slightly lower than test set loss. However, given a sufficient number of samples and the distribution of samples in the validation and test sets being similar, the validation loss was expected to be similar to the test set loss. The ECG windows from one individual were considered as the validation data and the remaining individuals were used for training the classifier. The classifier with the lowest error on the validation data was used to classify each window of the validation data as PTSD or control. The procedure was repeated for each individual in the data-set and the fraction of windows for each individual classified as PTSD was used to compute the AUROC curve. 
The overall training and evaluation of the deep neural networks is summarized in Table \ref{tbl:loocv}.

\subsubsection{Statistical tests}
A $\chi^2$ statistical test was performed to determine significant differences between AUROC curves.
Half of labels in the PTSD data-set were randomly assigned as PTSD and the remaining half as control before the LOOCV was performed. The AUROC curve was computed for each of the individuals and averaged. The process of randomly assigning labels and estimating the average AUROC curve was repeated 100 times to determine if the AUROC curve evaluated from the original data was statistically significantly higher than an AUROC curve 
evaluated 
from the lowest error on the validation data.


\begin{figure}[tbh]
     \centering
         \label{fig:y equals x}
     \caption{
     ROC curve for LOOT for each of the deep learning models and the baseline approach for classification of the PTSD data. Classification performance was expected to improve when different individuals formed the validation and training data. 
     }
     \label{fig:aucloocv}
\end{figure}

\section{Results}
\label{sec:reslt}
Table \ref{tbl:loocv} summarizes the PTSD classification performance 
of each model described in Section \ref{sssec:mdl}. 
Note that the highest performance scores are bolded.
Models 4 and 6 provided the best performance results in terms of AUC, F1 and BAcc. It is notable that the only consistent element in these models is the use of the artificial data and the PTSD data for training. However, when no real data was used, either for pre-training, or as a target, the results were poor. Model 2, which uses only artificial data simply classifies all data as `normal'. The ROC curves for LOOT with the PTSD data for each model is plotted in figure \ref{fig:aucloocv}. Classification performance achieved through training and testing model 6 proved accurate PTSD classification from ECG data could be achieved.


\begin{table}[tbh]
\centering
\caption{The AUROC curve (AUC), accuracy (Acc.), F1 score (F1) and balanced accuracy (BAcc.) for LOOT of models BL, 1, 2, 3, 4, 5 and 6 given in Section \ref{ssec:dnn}. `Arrhythmia data' and `Artificial data' 
refers to the real arrhythmia data from the 2021 PhysioNet challenge and the artificial TWA data respectively. Highest performances are bolded.
}
\begin{tabular}{|l|l|l|l|l|l|l|l|}
\hline
Model \# & Arrhythmia data & Artificial data & PTSD data & AUC & Acc. & F1 
& BAcc. \\ \hline
BL       & No & No & Yes & 0.56 & 0.67 & 0.57 & 0.67 \\ \hline
1        & No & No & Yes & 0.49 & 0.64 & 0.24 & 0.52\\ \hline
2        & No & Yes & No & 0.50 & 0.67 & 0.00 & 0.50\\ \hline
3        & Yes & No & Yes & 0.71 & 0.61 & 0.36 & 0.71\\ \hline
4        & No & \bf{Yes} & Yes & 0.74 & 0.69 & \bf{0.65} & \bf{0.73}\\ \hline
{5}        & {Yes} & \bf{Yes} & {No} & {0.63} & \bf{0.75} & {0.52} & {0.67} \\ \hline
6        & Yes & \bf{Yes} & Yes & \bf{0.77} & {0.72} & 0.64 & \bf{0.73} \\ \hline
\end{tabular}
\label{tbl:loocv}
\end{table}

\begin{table}[tbh]
\centering
\begin{tabular}{|l|l|l|l|}
\hline
Model \# & AUC $\pm$ $\sigma^2$  & Sensitivity $\pm$ $\sigma^2$ & Specificity $\pm$ $\sigma^2$ \\ \hline
BL      & 0.45 $\pm$ 0.001 & 0.39 $\pm$ 0.002 & 0.93 $\pm$ 0.00005 \\ \hline
1       & 0.62 $\pm$ 0.002 & 0.79 $\pm$ 0.004 & 0.43 $\pm$ 0.008 \\ \hline
2       & 1 & 1 & 1 \\ \hline
3       & 0.60 $\pm$ 0.0002 & 0.71 $\pm$ 0.02 & 0.45 $\pm$ 0.02 \\ \hline
4       & 0.85 $\pm$ 0.01 & 0.71 $\pm$ 0.04 & {0.89 $\pm$ 0.005}\\ \hline
{5}     & {1} & {1} & {1}  \\ \hline
6       & {0.98 $\pm$ 0.003} & {0.93 $\pm$ 0.002} & 0.93 $\pm$ 0.0006  \\ \hline
\end{tabular}
\caption{The mean and variance ($\sigma^2$) for the AUROC curve (AUC), sensitivity and specificity of classifying the artificial TWA data using LOOT are reported for the models given in section \ref{sssec:mdl}. For models 2 and 5, not trained on the PTSD data, the classification metrics for the test fold of the artificial TWA data are reported.}
\label{tbl:artif_twa}
\end{table}

Table \ref{tbl:artif_twa} summarizes classification results for the completely trained models described in section \ref{sssec:mdl} on the artificial TWA data
. Models 1 and 3 achieved average AUROC curves of 0.62 and 0.60, respectively, for classifying the artificial TWA data. Model 3 achieved a sensitivity of 0.71, suggesting identifying TWAs was an important feature learned from the PTSD data for accurate classification. Models 4 and 6 classified the artificial TWA data with average AUROC curves of 0.85 and 0.98 respectively. Model 6 particularly classified the artificial TWA data accurately, suggesting after the transfer learning on the PTSD data, TWA detection was important for PTSD classification. The $\chi^2$ test yielded the output from each trained model 6 was independent of the true label. However, given the large number of ECGs in the artificial TWA data-set, the test was over-powered, clearly indicated by the average AUROC of 0.98 with a small variance in table \ref{tbl:artif_twa}.


\section{Discussion}
The results in table \ref{tbl:loocv} clearly indicate that pre-training on a large database of ECGs (over 70,000) with varying arrhythmias and other clinical abnormalities (Model 3), or a large artificial database (of 180,000) ECGs (Model 4), provides a significant boost in performance over both standard and DNN-based baseline approaches (model BL, and Model 1) for the chosen classification problem. Specifically, in the case of pre-training on real arrhythmia data alone, the AUC rose from 0.56 to 0.71, and to 0.74 for pre-training on artificial TWA data alone. Pre-training on artificial data led to a boost for all metrics, providing the highest F1 and BAcc. (Using real arrhythmia data only led to a drop in Acc. and F1 over the BL model.) Pre-training on both the real arrhythmia data and artificial data led to the largest increase in AUC, although it led to inferior results (in F1 and BAcc) compared to excluding the real arrhythmia data. Importantly, this indicates that the artificial data was the key component in boosting performance.

It is interesting to note that when the real PTSD data was not used (Model 2), significant performance reduction was observed, as expected, even though the artificial data that was tuned to exhibit similar characteristics to the PTSD cohort. This indicates that the artificial data does not closely match the distributions of the real target data. Therefore, the artificial data is likely to be providing a coarse tuning of the deep neural network on a very broad range of simulated TWA data, allowing the network to focus in on  features related to PTSD. In addition, pre-training on real arrhythmia data leads to a network that recognizes and differentiates between rhythms, although this somewhat biases the classifier towards predicting patients are `normal'. 
This is consistent with the literature, where TWAs and arrhythmias have been shown to be important markers of PTSD, but with varying specificity \cite{Vaccarino2013ptsd,LAMPERT20151000}.

It is also important to note that when the DNN trained on all data using transfer learning (model 6) is back-tested on the artificial TWA data, the performance drops only slightly (from an AUC of 1.0 to 0.98 - see table \ref{tbl:artif_twa}). This provides evidence that using artificial data does not cause a loss of generalization in the model even though the final target data included only 36 subjects.

\section{Conclusion}
The work presented in this article demonstrates the utility of a realistic model to significantly boost training and test performance on a small dataset. Our results indicate that the model was the single most important part of the transfer learning process, boosting performance by more than either the source (arrhythmia) data, or the target (PTSD) data. 

This result is significant for several reasons. 
First, in the biological sciences, and healthcare in particular, it is common to be resource limited, and only have a small collection of high-quality data. Large volume collection can be prohibitive because of costs, legal/privacy barriers, social resistance to data acquisition, or the remoteness of the population.  Secondly, many diseases are quite rare, and it is impracticable to assemble large databases amenable to machine learning. 
Thirdly, even if a large database can be collected, they are typically difficult to curate, and often the quality of the labels drops as the volume increases. Fourthly, large databases tend to disadvantage under-represented minorities, or individuals from resource-constrained areas such as the rural US, or LMICs in general. 

It is interesting to consider that the principle of data augmentation using artificial data could be applied to similar problems to improve classification performance on the target domain data, provided the model used for data generation accurately captured the features characteristic of the target domain data.  Finally, we note that because the model employed is computationally efficient, the artificial database can be created on-the-fly, and stored in memory, thus reducing storage costs if needed.


We note a key limitation of the study is that \emph{because} our target dataset is relatively small and imbalanced (only 36 individuals), further work is required to identify if the proposed methodology will generalize beyond the population under consideration. (However, the back-testing on artificial data presents some evidence to support the idea of generalization.) To address this, we are developing similar approaches on other cardiac conditions where larger numbers of target data are available. By subsampling such data, we can test the idea presented in this work.   

It is interesting to consider that the principle of data augmentation using artificial data could be applied to similar problems to improve classification performance on the target domain data, provided the model used for data generation accurately captures the features characteristic of the target domain data.  Finally, we note that because the model employed is computationally efficient, the artificial database can be created on-the-fly, and stored in memory, thus reducing storage costs if needed.
Given that data sourcing, transfer and preservation are costly, and compute is relatively cheap, our approach could have enormous potential in the biological sciences. Moreover, the burden and ethical considerations of collection of data from humans and animals for medical research are considerable, and increasingly the focus of attention. Ideally, approaches such as the ones presented here might usher in a new era of in-silico experimentation in a manner akin to the switch over to nuclear weapons testing in the 1990's by most countries.

\bibliographystyle{unsrt}
\bibliography{references}
\end{document}